\documentclass[runningheads]{llncs}
\usepackage[T1]{fontenc}
\usepackage{graphicx}

\usepackage{amsmath,amsfonts}
\usepackage{algorithm}
\usepackage{array}
\usepackage[caption=false,font=normalsize,labelfont=sf,textfont=sf]{subfig}
\usepackage{textcomp}
\usepackage{stfloats}
\usepackage{url}
\usepackage{verbatim}
\usepackage{graphicx}
\usepackage{float}
\usepackage{bm}
\usepackage{multirow}
\usepackage{xcolor}
\usepackage{algpseudocode}
\usepackage{cite}

\usepackage{wrapfig}
\usepackage[misc]{ifsym} % Letter for corresponding author

\algnewcommand\algorithmicparam{\textbf{Parameter:}}
\algnewcommand\Param{\item[\algorithmicparam]}
\graphicspath{ {./figure/} }
\begin{document}
%
% \title{Contribution Title\thanks{Supported by organization x.}}

\title{AFLOW: Developing Adversarial Examples under Extremely Noise-limited Settings\thanks{This paper was accepted by ICICS 2023.}}

\titlerunning{AFLOW}
\author{Renyang Liu\inst{1}\orcidID{0000-0002-7121-1257} \and
Jinhong Zhang\inst{2}\orcidID{0000-0002-9906-3508} \and
Haoran Li\inst{2}\orcidID{0000-0002-0409-2227} \and
Jin Zhang\inst{3}\orcidID{0009-0007-8545-8203} \and
Yuanyu Wang\inst{3}\orcidID{0009-0001-8595-1168} \and
Wei Zhou\inst{\textsuperscript{2,~\Letter}}  \orcidID{0000-0002-5881-9436}}
\authorrunning{Renyang Liu et al.}

\institute{School of Information Science and Engineering, Yunnan University, Kunming, China\\
\email{ryliu@mail.ynu.edu.cn}\\ 
\and
Engineering Research Center of Cyberspace, Yunnan University, Kunming, China \\
\email{\{jhnova,lihaoran\}@mail.ynu.edu.cn,zwei@ynu.edu.cn} \\
\and
Kunming Institute of Physics, Kunming, China \\
\email{\{zhangjin\_211,wxyjin232425\}@163.com}}

\maketitle              % typeset the header of the contribution
\begin{abstract}
Extensive studies have demonstrated that deep neural networks (DNNs) are vulnerable to adversarial attacks. Despite the significant progress in the attack success rate that has been made recently, the adversarial noise generated by most of the existing attack methods is still too conspicuous to the human eyes and proved to be easily detected by defense mechanisms. Resulting that these malicious examples cannot contribute to exploring the vulnerabilities of existing DNNs sufficiently. Thus, to better reveal the defects of DNNs and further help enhance their robustness under noise-limited situations, a new inconspicuous adversarial examples generation method is exactly needed to be proposed. To bridge this gap, we propose a novel Normalize Flow-based end-to-end attack framework, called AFLOW, to synthesize imperceptible adversarial examples under strict constraints. Specifically, rather than the noise-adding manner, AFLOW directly perturbs the hidden representation of the corresponding image to craft the desired adversarial examples. Compared with existing methods, extensive experiments on three benchmark datasets show that the adversarial examples built by AFLOW exhibit superiority in imperceptibility, image quality and attack capability. Even on robust models, AFLOW can still achieve higher attack results than previous methods.

\end{abstract}
\section{Introduction}
Deep Neural Networks (DNNs) have shown their excellent performance in a wide variety of deep learning tasks, such as Computer Vision (CV) \cite{DBLP:journals/scn/WangLWMSW22}, Natural Language Processing (NLP) \cite{DBLP:conf/wsdm/WuW0ZW22}, and Autonomous Driving \cite{DBLP:journals/tits/KiranSTMSYP22}. However, 
\begin{wrapfigure}{r}{6cm}
% \vspace{-20pt}
\centering
    \includegraphics[width=0.49\textwidth]{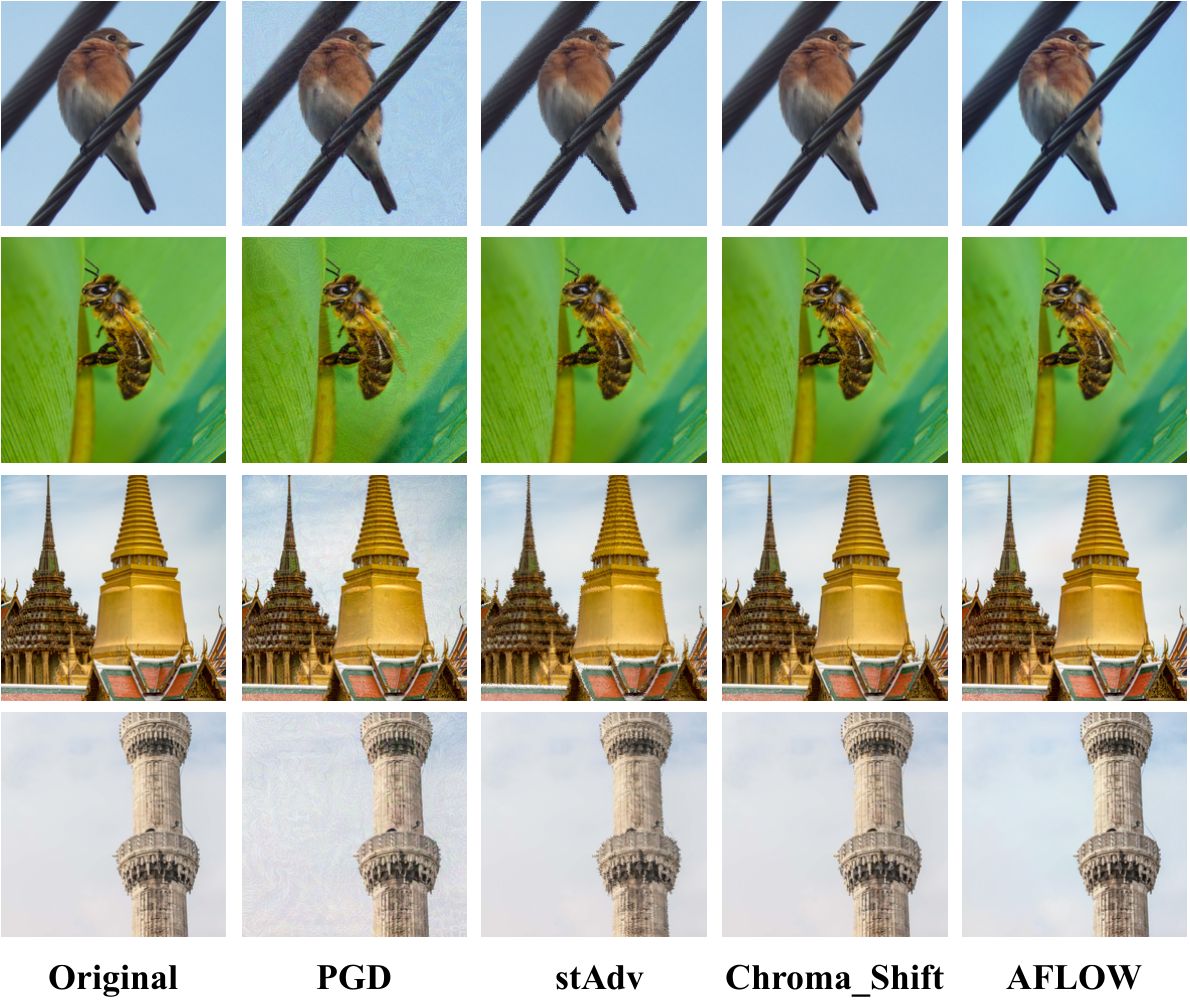}
    \caption{The original images and the adversarial examples generated by PGD \cite{DBLP:conf/iclr/MadryMSTV18}, 
    stAdv \cite{DBLP:conf/iclr/XiaoZ0HLS18}, Chroma-Shift \cite{DBLP:conf/mm/AydinSKHT21} and the proposed AFLOW for the ResNet-152 \cite{DBLP:conf/cvpr/HeZRS16} model.}
    \label{fig:adversarial}    
    \vspace{-0.8cm}
\end{wrapfigure}
the DNNs have been demonstrated to be vulnerable to adversarial examples \cite{DBLP:journals/corr/SzegedyZSBEGF13}, especially in CV, which usually build by adding elaborate well-designed noise to the original clean image. Typically, the adversarial examples should have the following two characteristics: One is the attack ability, which means that the adversarial examples can fool the well-trained DNN models to output the wrong predictions; the other is the imperceptibility, which means the added noise is unnoticeable to human eyes.

Recently, researchers have carried out many studies on adversarial examples, including adversarial attack approaches and their corresponding defense techniques. In CV, existing attack methods usually generate adversarial examples by optimizing noise and adding them to the benign image \cite{DBLP:journals/corr/GoodfellowSS14, DBLP:conf/iclr/IlyasEM19, DBLP:journals/compsec/PengLWCWCZ21,DBLP:journals/tdsc/HeWLYJLZ23,DBLP:conf/iclr/ChenMSG0LF22}, and achieved admirable attack ability. However, these methods ignore another critical characteristic, which constrains the perturbation of a liberal policy. Most methods only consider the $L_p$-norm as a condition to ensure that the perturbation is unnoticeable, e.g., $L_{inf}=\{8,16,32,64\}$, which is the max difference value between the clean image and evil image. While the $L_p$-norm is not enough to preserve the vivid details of the generated adversarial examples, resulting in apparent adversarial noise. Some pioneer works make a step forward on inconspicuous attacks, like stAdv \cite{DBLP:conf/iclr/XiaoZ0HLS18}, Chroma-Shift \cite{DBLP:conf/mm/AydinSKHT21}, and FIA \cite{DBLP:conf/cvpr/LuoL0WXS22} build evil examples by spatial transform techniques or by manipulating the image in the frequency level rather than in a noise-adding way. However, the generated evil image still carries many burrs; thus, it can be easily detected \cite{DBLP:conf/sp/LingJZWWLW19, DBLP:conf/cvpr/LiDLL20, DBLP:conf/eccv/LiXQHLZ22,DBLP:conf/atal/ChenGZLL21}, which is infaust for further study of the susceptibility of DNNs and improving the existing DNNs' robustness.

Notably, rare research has been proposed to explore the vulnerability and robustness of DNNs for adversarial examples built under rigorous constraints. In this regard, designing a method to generate more inconspicuous adversarial examples under strict constraints is essential to AI applications. It can make a huge step forward in sufficiently exploring the fragility and guiding the robustness improvement of the existing DNNs. In addition, the crafted adversarial noise should be more invisible and challenging to be detected by the defense mechanism. 

To bridge this gap, in this paper, we intend to generate adversarial examples in the rigorous noise-limited scenario to explore the vulnerability of existing DNNs. The noise-limited setting means that the $L_{inf}$-norm of the generated adversarial perturbation is strictly restricted, which is beneficial to improve the imperceptibility of calculated adversarial perturbations and preserve the image quality of the generated adversarial examples as well. In order to balance the invisibility and the attack ability of the generated adversarial examples, a novel Normalize Flow (NF) model \cite{DBLP:journals/ijautcomp/XuMLDLTJ20} based attack method called AFLOW, has been proposed to deal with the issues mentioned above. Benefiting from the splendid reconstruction capability of the NF model, we can generate adversarial examples by slightly disturbing the hidden space of the clean images. Specifically, the AFLOW first input the clean image $x$ into the well-trained NF model to obtain its hidden representation $z_0$. Next, we regard the $z_0$ as the initial point and optimize it to $z_t$ until it has reversed $x_t$ can attack the target model successfully. Empirically, the proposed AFLOW can significantly preserve the generated adversarial examples' image quality while achieving an admirable attack success rate.

We conduct extensive experiments on three different computer vision benchmark datasets. In strict noise-limited scenarios, empirical results show that the AFLOW can craft adversarial examples with better invisibility and excellent image quality while achieving a remarkable attack performance. As shown in Fig. \ref{fig:adversarial}, comparing with the existing methods, such as PGD \cite{DBLP:conf/iclr/MadryMSTV18}, stAdv \cite{DBLP:conf/iclr/XiaoZ0HLS18}, and Chroma-Shift \cite{DBLP:conf/mm/AydinSKHT21}, the adversarial examples generated by AFLOW is indistinguishable from the original images. The main contributions of this work could be summarized as follows:

\begin{itemize}
    \item We tried to improve the detection resistance and attack performance under rigorous noise-limited settings due to the adversarial examples crafted by existing attack methods that can be easily detected by adversarial detectors. Moreover, in this situation, the attack performance of existing methods have been faded significantly.    
  
    \item we design a novel end-to-end scheme called AFLOW to craft adversarial examples for noise-limited settings by directly disturbing the latent representation of the clean examples rather than noise-adding. This method can generate adversarial examples with high attack performance and imperceptibility.
    
    \item We conduct comprehensive experiments on three real-world datasets, and the results demonstrate the superiority of AFLOW in synthesizing adversarial examples under noise-limited attack settings. Compared to existing baselines, the adversarial examples built by AFLOW have high attack ability, outstanding invisibility and excellent image quality. Notably, AFLOW achieves up to 96.73\% ASR under the constraint is $L_{inf}=1$ on the ImageNet dataset.
\end{itemize}

The rest of this paper is organized as follows. We first briefly review the methods relating to imperceptible adversarial attacks in Sec. \ref{Sec:related}. Then, Sec. \ref{sec:methodology} introduces the details of the proposed AFLOW framework. Finally, the experiments are presented in Sec. \ref{Sec:experiments}, with the conclusion drawn in Sec. \ref{Sec:conclusion}.

\section{Related Work}
\label{Sec:related}
In this section, we briefly review the most pertinent attack methods to the proposed work. The adversarial attacks and the techniques used for crafting inconspicuous adversarial perturbations.

\subsection{Adversarial Attack}
The adversarial attack has already been intensely investigated in recent years. Szegedy et al. demonstrated that it was possible to mislead the deep neural networks (DNNs) by adding imperceptible and well-designed perturbations to the original benign input image. They simplified the problem of generating adversarial examples by disturbing the loss function by a small margin, which was then solved by L-BFGS \cite{DBLP:journals/corr/SzegedyZSBEGF13}. Goodfellow et al. proposed an effective un-targeted attack method called Fast Gradient Sign Method (FGSM) \cite{DBLP:journals/corr/GoodfellowSS14}, which generated adversarial examples under the $L_\infty$ norm limit of the perturbation. Kurakin et al. proposed the BIM \cite{DBLP:conf/iclr/KurakinGB17a}, which executed FGSM iteratively with a small update step in each epoch, to ensure that the update direction of gradients could be more accurate. Projected gradient descent (PGD) \cite{DBLP:conf/iclr/MadryMSTV18} could be regarded as a generalized version of BIM. Inspired by momentum, Dong et al. \cite{DBLP:conf/cvpr/DongLPS0HL18} proposed Momentum Iterative FGSM (MI-FGSM), which integrated momentum into the iterative BIM process. Like L-BFGS, Carlini and Wagner proposed a set of optimized adversarial attack C\&W \cite{DBLP:conf/sp/Carlini017} to craft adversarial examples under the limit of $ L_0 $, $ L_2 $, and $ L_\infty $ norm. 

\subsection{Imperceptible Adversarial Attacks}
Unlike the previous methods, which synthesize adversarial examples by adding noise and then clipping the adversarial examples use $L_p$-norm based metrics to ensure the adversarial examples' invisibility. Xiao et al. propose a spatial transform-based (flow field) method, stAdv \cite{DBLP:conf/iclr/XiaoZ0HLS18}, to generate adversarial examples. This approach is based on altering the pixel positions rather than modifying the pixel value and brings a booming prospect that the DNNs can be fooled only by pixel shifts and make a step forward to explore vulnerability more deeply. Chroma-shift \cite{DBLP:conf/mm/AydinSKHT21}, which calculates the flow field in the image's YUV space rather than RGB space, make another step forward to fabricate adversarial examples with higher human imperceptibility. Besides, Adv\_Cam \cite{DBLP:conf/cvpr/DuanM00QY20} adopt style transfer techniques to generate adversarial images more natural for the physical world. 

The most related method to the current work is AdvFlow \cite{DBLP:conf/nips/DolatabadiEL20}, which uses the Normalizing Flow model to map the input image to a hidden representation $z$. And then adding an optimized noise $\mu$ to $z$ to generate the representation of the corresponding adversarial example. Note that AdvFlow is designed for black-box settings and generates adversarial examples in a noise-adding and limitation way, which requires many queries to perform a successful attack.

Therefore, generating inconspicuous adversarial examples poses the request for a method that can craft adversarial examples with strong attack ability, high imperceptibility, and high image quality. Besides, the attack strategy must be direct, efficient, and effective to perform attacks for different models and datasets. To achieve this goal, we know from the previous studies that the Normalize Flow model can transform an image between pixel space and hidden space. Besides, disturbing images in their hidden representations can convert to an adversarial example at the pixel level. This could help us to explore existing models' vulnerabilities under rigour noise constraints. Hence, we are well motivated to develop a Normalize Flow-based scheme to generate adversarial examples with better human visual perception.

\section{Methodology}
\label{sec:methodology}
In this section, we propose our attack method. First, we take an overview of our method. Next, we go over the detail of each part step by step. Finally, we discuss our objective function and summarize the whole process as Alg. \ref{alg:alg1}. 

\subsection{Overview}
\begin{figure}[!htp]
      \centering
      \includegraphics[width=\textwidth]{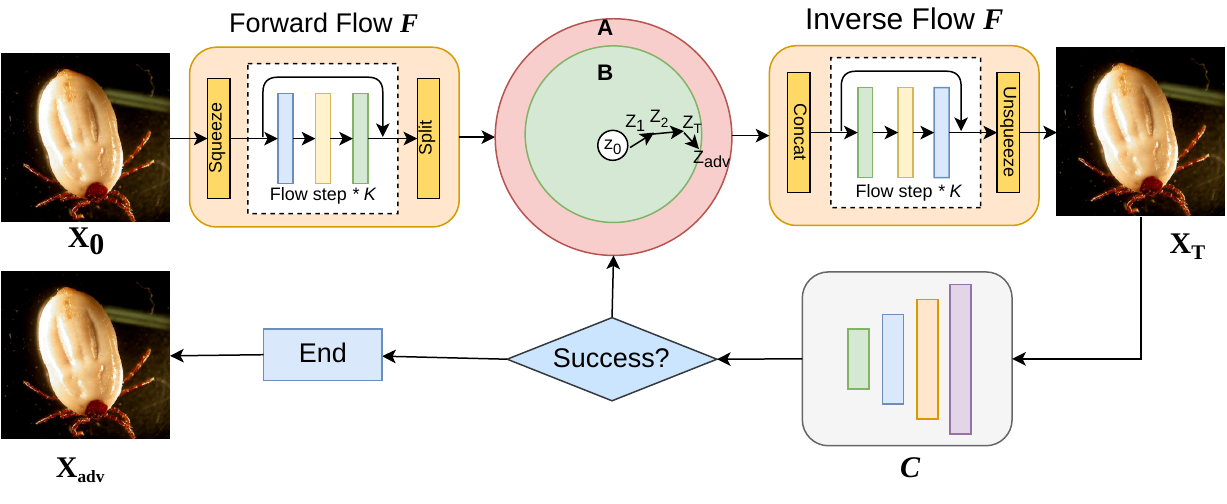}
      % \vspace{-0.9 cm}
      \caption{The framework of proposed AFLOW. $\bm{X}$ represent the image, among them, $\bm{X_0}$ is the benign image, $\bm{X_T}$ is the intermediate results and $\bm{X_{adv}}$ is the corresponding adversarial counterpart; $Z$ is the hidden representation of the image; among them, the $Z_0$ is the benign hidden value, $Z_1 \sim Z_T$ are the intermediate results, and the $Z_{adv}$ is the adversarial hidden value; $\bm{A}$ represents the adversarial space and $\bm{B}$ is the benign space; $\bm{F}$ is the well-trained Normalize Flow model and $\bm{C}$ is the pre-trained classifier.}
      \label{fig:framwork}
\end{figure}

The proposed AFLOW attack framework can be divided into three parts, the first one is to map clean image $x$ to its latent space $z$, which we are going to make changes, and the second part is to disturb $z$ to $z_T$ in an iterative manner; the last one is doing the inverse operation to translate $z_T$ to its corresponding RGB space counterpart, that is, the candidate adversarial example $X_T$ until it can fool the target DNN model to make wrong decisions. The whole process is shown in Fig. \ref{fig:framwork}.

\subsection{Problem Statement}
Given a well-trained DNN classifier $ \mathcal{C} $ and a correctly classified input $ (x,y) \sim D $, we have $ 
 \mathcal{C}(x)=y $, where $D$ denotes the accessible dataset. The adversarial example $ x_{adv} $ is a neighbor of $ x $ and satisfies that $ \mathcal{C}(x_{adv}) \neq y$ and $ \left \| x_{adv}-x \right \|_p \leq \epsilon $, where the $L_p$ norm is used as the metric function and $ \epsilon $ is usually a small noise budget. With this definition, the problem of finding an adversarial example becomes a constrained optimization problem:

\begin{equation}
    x_{adv}= \begin{cases}
    \underset{\left \| x_{adv}-x \right \|_p \leq \epsilon}{arg\ max\ \mathcal{L}} ( \mathcal{C}(x_{adv}) \neq y),\quad & un-targeted \\
    \underset{\left \| x_{adv}-x \right \|_p \leq \epsilon}{arg\ min\ \mathcal{L}} ( \mathcal{C}(x_{adv}) = t),\quad & targeted
    \end{cases} 
\end{equation}
where $ \mathcal{L} $ stands for a loss function that measures the confidence of the model outputs, and $t$ is the target label.

\subsection{Normalizing Flow}
Normalizing Flows (NF) \cite{DBLP:journals/ijautcomp/XuMLDLTJ20} are a class of probabilistic generative models, which are constructed based on a series of completely reversible components. The reversible property allows to transform from the original distribution to a new one and vice versa. By optimizing the model, a simple distribution (such as the Gaussian distribution) can be transformed into a complex distribution of real data. The training process of normalizing flows is indeed an explicit likelihood maximization. Considering that the model is expressed by a fully invertible and differentiable function that transfers a random vector $ \bm{z} $ from the Gaussian distribution to another vector $ \bm{x} $, we can employ such a model to generate high dimensional and complex data.

Specifically, given a reversible function $ f:\mathbb{R}^d\rightarrow \mathbb{R}^d $ and two random variables $ z\sim p(z) $ and $ z'\sim p(z') $ where $ z' = f(z) $, the change of variable rule tells that
\begin{equation}
      p(z')=p(z)\left | det \frac{\partial {f^{-1}}}{\partial{z'}} \right |, \quad
      p(z)=p(z')\left | det \frac{\partial {f}}{\partial{z}} \right |      
\end{equation}
where $ det $ denotes the determinant operation. The above equation follows a chaining rule, in which a series of invertible mappings can be chained to approximate a sufficiently complex distribution, i.e.,
\begin{equation}
      \label{eq:eq2}
      z_K = f_K \odot ... \odot  f_2 \odot 
f_1(z_0),
\end{equation}
where each $ f $ is a reversible function called a flow step. Eq. \ref{eq:eq2} is the shorthand of $ f_K(f_{k-1}(...f_1(x))) $. Assuming that $x$ is the observed example and $z$ is the hidden representation, we write the generative process as
\begin{equation}
      x=f_{\theta}(z),
\end{equation}
where $ f_{\theta} $ is the accumulate sum of all $ f $ in Eq. \ref{eq:eq2}. Based on the change-of-variables theorem, we write the log-density function of $ x=z_K $ as follows:
\begin{equation}
      \label{eq:eq21}
      -\log{p_K}(z_K)=-\log p_0(z_0)-\sum_{k=1}^{K}\log\left | det\frac{\partial z_{k-1}}{\partial z_{k}} \right |,
\end{equation}
where we use $z_k=f_k(z_{k-1})$ implicitly. The training process of normalizing flow minimizes the above function, which exactly maximizes the likelihood of the observed training data. Hence, the optimization is stable and easy to implement. 

\begin{algorithm}[htp]
	\caption{Normalizing Flow-based Spatial Transform Attack}
	\label{alg:alg1}
	\begin{algorithmic}[1]
		\Require
		$ X_{tr} $: a batch of clean examples used for training;  $ \alpha $: the learning rate; $ T $: the maximal training iterations; $ Q $: the maximal querying number; $ \epsilon $: the noise budget; $ X_{te} $: a clean example used for test; $ \mathcal{C} $: the target model to be attacked.
		
		\Ensure
		The adversarial example $x_{adv}$ is used for attack.
		
		\Param
		The flow model $f_\theta$.
		
		\State Initialize the parameters of the flow model $f_\theta$;
		\For{$i=1$ to $T$}
		\State Optimize $f_\theta$ according to Eq. \ref{eq:eq21};
		\If{Convergence reached}
		\State break;
		\EndIf
		\EndFor
		\State Obtain optimized $f_\theta$;
		\State Compute the hidden representation of examples in $X_{te}$ via $ z=f^{-1}(x_{te})$;
	    \State $z_{0}^{'}=z$
		\For {$i=1$ to $Q$}
		\State Optimize $z_{i}^{'}$ via Eq. \ref{eq:loss};
		\State Compute the adversarial example candidate $x_{i}^{'}$ via  $x’=f(z_{i}^{'})$;
		\State Clip the example via $ Clip() $;
		\If {Successfully attack $\mathcal{C}$ by $x_{i}^{'}$}
		\State $x_{adv} = x_{i}^{'}$
		\State break.
		\EndIf
		\EndFor
	\end{algorithmic}
\end{algorithm}

\subsection{Generation of Adversarial Examples}
Given a well-trained flow model $f_\theta$ and a normal input $x$, to generate an adversarial example, we first calculate its corresponding latent space vector $z$ by performing a forward flow process via $z = f_{\theta}(x)$. Once the $z$ is calculated, we regard $z$ as the perturbation starting point of the latent adversarial $z'$, then directly optimize it with the Adam optimizer, and finally restore the optimized $z'$ to the image space through the inverse operation of the Normalizing Flow model, that is $ x'=f_{\theta} (z') $, to get its perturbed example $x'$ in pixel level. We will repeat the above process to optimize $z'$ until $x'$ becomes an eligible adversarial example. For the fairness of comparison, we follow the existing attack methods which constrain the perturbation within a certain range. Once we obtain the adversarial example candidate $x'$, we employ the clip function $x'=x'+Clip(-\epsilon, x'-x, \epsilon)$ to ensure the imperceptible property of the perturbation, where $\epsilon$ is the acceptable noise budget, in this paper, $ \epsilon \in {1, 2, 4, 8}$.

\subsection{Objective Functions}
In order to take into account the attack success rate and visual invisibility of the generated adversarial examples, which keeps it as similar as possible to the benign image to ensure that it is imperceptible to human eyes. For adversarial attacks, the goal is making $\mathcal{C}(X_{adv}) \neq y$, we give the objective function as:

\begin{equation}
    \label{eq:loss}
    \begin{cases}
    \mathcal{L}_{adv} (X,y)= max[\mathcal{C}(X_{adv})_{y}-\underset{k\neq y}{max}\mathcal{C}(X_{adv})_{k},k],\quad & un-targeted \\
    \mathcal{L}_{adv} (X,y,t)= min[\underset{k=t}{max}\mathcal{C}(X_{adv})_{k} - \mathcal{C}(X_{adv})_{y},k],\quad & targeted
    \end{cases} 
\end{equation}

The whole algorithm of AFLOW is listed in Alg. \ref{alg:alg1}, which could help readers to re-implement our method step-by-step.

\section{Experiments}
\label{Sec:experiments}
In this section, we evaluate the proposed AFLOW on three benchmark image classification datasets. We first compare our proposed method with several baseline techniques concerned with Attack Success Rate (ASR) on clean models and robust models on three CV baseline datasets under strong constraints. Then, we evaluate the anti-detection ability of the proposed and baseline methods. Finally, we first provide a comparative experiment to the existing attack methods in image quality or similarity aspects with regard to LPIPS, DISTS, SSIM, and PSNR et. al.. Through these experimental results, we show the superiority of our method in attack ability, human perception, and image quality.

\begin{table}[]
\caption{Experimental results on the attack success rate of \textbf{un-targeted} attack on dataset Caltech256 under $l_{inf} $ noise budget is 1, 2, and 4, respectively.}
\label{tab:Caltech256}
\setlength\tabcolsep{2pt}
\renewcommand{\arraystretch}{1.1}
\centering
\resizebox{\linewidth}{!}{
\begin{tabular}{c|c|ccccccccc}
\hline
Epsilon            & Model        & BIM            & PGD   & MIFGSM & TIFGSM & DIFGSM & APGD  & Jitter & AdvFlow & AFLOW           \\ \hline
\multirow{4}{*}{1} & VGG-19       & 31.35          & 35.64 & 40.72  & 2.93   & 26.66  & 27.66 & 16.89  & 0.58    & \textbf{82.81}  \\
                   & ResNet-152   & 37.79          & 41.11 & 51.17  & 6.35   & 28.42  & 40.66 & 26.17  & 1.75    & \textbf{88.67}  \\
                   & MobileNetV2  & 48.97          & 46.38 & 59.15  & 7.76   & 32.28  & 36.83 & 26.63  & 3.73    & \textbf{91.02}  \\
                   & ShuffleNetV2 & 63.75          & 65.49 & 71.46  & 17.22  & 47.11  & 23.38 & 48.90  & 16.67   & \textbf{88.67}  \\ \hline
\multirow{4}{*}{2} & VGG-19       & 79.59          & 83.50 & 82.91  & 25.29  & 78.42  & 57.71 & 61.91  & 5.13    & \textbf{97.27}  \\
                   & ResNet-152   & 87.01          & 87.30 & 86.50  & 30.18  & 77.83  & 73.14 & 66.86  & 11.28   & \textbf{98.83}  \\
                   & MobileNetV2  & 88.74          & 93.68 & 89.73  & 38.33  & 86.92  & 67.72 & 68.01  & 21.64   & \textbf{99.22}  \\
                   & ShuffleNetV2 & 93.89          & 93.00 & 94.13  & 36.43  & 85.19  & 31.84 & 69.02  & 33.08   & \textbf{97.27}  \\ \hline
\multirow{4}{*}{4} & VGG-19       & 97.46          & 99.12 & 97.65  & 75.29  & 97.95  & 66.70 & 86.41  & 32.82   & \textbf{99.61}  \\
                   & ResNet-152   & 97.07          & 98.54 & 97.07  & 70.31  & 98.34  & 81.25 & 89.36  & 44.19   & \textbf{99.61}  \\
                   & MobileNetV2  & 99.11          & 99.31 & 97.92  & 83.35  & 99.41  & 69.63 & 90.92  & 50.78   & \textbf{100.00} \\
                   & ShuffleNetV2 & \textbf{99.90} & 99.71 & 99.01  & 75.56  & 99.01  & 33.73 & 83.65  & 55.47   & 99.61  \\ \hline
\end{tabular}
}
\end{table}

% \vspace{-25pt}
\subsection{Settings}
\textbf{Dataset:} We verify the performance of our method on three benchmark datasets for the computer vision task, named Caltech-256\footnote{https://data.caltech.edu/records/nyy15-4j048} \cite{griffin2007caltech}, ImageNet-1k\footnote{https://image-net.org/} \cite{DBLP:conf/cvpr/DengDSLL009} and Places365\footnote{http://places2.csail.mit.edu/index.html} \cite{zhou2017places}. In detail, the Caltech256 dataset consists of 30,607 real-world images of different sizes, spanning 257 classes (256 object classes and an additional clutter class). ImageNet-1K has 1,000 categories, containing about 1.3M examples for training and 50,000 examples for validation. The places365 is composed of 10 million images comprising 434 scene classes. 

In particular, in this paper, we extend our attack on the whole images of Caltech256. And for ImageNet-1K, we carry out our attack on its subset datasets from the NIPS2017 Adversarial Learning Challenge, and we call it NIPS2017 in the later chapters. Regarding the Places365 dataset, we use its val\_256 subset for all the experiments.

% \vspace{-15pt}
\begin{table}[]
\caption{Experimental results on the attack success rate of \textbf{un-targeted} attack on dataset Places365 under $l_{inf} $ noise budget is 1, 2, and 4, respectively.}
\label{tab:Places365}
\setlength\tabcolsep{2pt}
\renewcommand{\arraystretch}{1.2}
\centering
\resizebox{\linewidth}{!}{
\begin{tabular}{c|c|ccccccccc}
\hline
Epsilon            & Model        & BIM   & PGD   & MIFGSM & TIFGSM & DIFGSM & APGD  & Jitter & AdvFlow & AFLOW           \\ \hline
\multirow{4}{*}{1} & VGG-19       & 41.43 & 44.59 & 52.29  & 8.15   & 32.20  & 12.98 & 17.99  & 9.3     & \textbf{98.05}  \\
                   & ResNet-152   & 33.43 & 37.71 & 49.65  & 6.90   & 28.22  & 12.44 & 16.19  & 17.69   & \textbf{99.61}  \\
                   & MobileNetV2  & 52.81 & 55.30 & 65.22  & 17.00  & 40.84  & 13.09 & 27.48  & 31.54   & \textbf{99.61}  \\
                   & ShuffleNetV2 & 68.96 & 69.92 & 78.40  & 21.86  & 52.13  & 5.78  & 34.03  & 47.29   & \textbf{96.88}  \\ \hline
\multirow{4}{*}{2} & VGG-19       & 88.62 & 87.15 & 92.00  & 39.22  & 84.52  & 24.61 & 57.24  & 32.35   & \textbf{100.00} \\
                   & ResNet-152   & 82.60 & 84.49 & 87.46  & 33.23  & 74.75  & 23.44 & 51.83  & 62.02   & \textbf{99.61}  \\
                   & MobileNetV2  & 92.59 & 92.44 & 92.94  & 53.97  & 87.92  & 23.99 & 58.07  & 54.62   & \textbf{100.00} \\
                   & ShuffleNetV2 & 95.21 & 95.01 & 94.74  & 46.35  & 88.71  & 10.67 & 58.96  & 55.47   & \textbf{100.00} \\ \hline
\multirow{4}{*}{4} & VGG-19       & 98.91 & 99.51 & 99.03  & 82.01  & 99.12  & 27.15 & 84.77  & 71.09   & \textbf{100.00} \\
                   & ResNet-152   & 98.02 & 98.51 & 98.22  & 76.82  & 98.02  & 27.73 & 79.96  & 87.5    & \textbf{100.00} \\
                   & MobileNetV2  & 99.50 & 99.32 & 98.82  & 88.52  & 99.60  & 26.82 & 83.76  & 91.41   & \textbf{100.00} \\
                   & ShuffleNetV2 & 99.21 & 99.70 & 99.60  & 82.27  & 99.30  & 11.66 & 74.58  & 84.38   & \textbf{100.00} \\ \hline
\end{tabular}
}
\end{table}

% \vspace{-15pt}

\textbf{Models:} 
For NIPS2017, we use the PyTorch pre-trained clean model VGG-19 \cite{DBLP:journals/corr/SimonyanZ14a}, ResNet-152 \cite{DBLP:conf/cvpr/HeZRS16}, MobileNet-V2 \cite{DBLP:journals/corr/abs-1801-04381} and ShuffleNet-V2 \cite{DBLP:conf/eccv/MaZZS18} as the victim models. For Caltech256 and Places365, we utilize the transfer learning to train the ImageNet pre-trained VGG-19, ResNet-152, MobileNet-V2 and ShuffleNet-V2, with top-1 classification accuracy 93.65\%, 98.43\%, 96.21\%, 73.85\% on Caltech256 and 96.63\%, 98.64\%, 79.71\%, 65.89\% on Places365, respectively.

And in terms of robust models, they are including Salman2020Do\_R50\cite{DBLP:conf/nips/SalmanIEKM20}, Salman2020Do\_R18\cite{DBLP:conf/nips/SalmanIEKM20}, Engstrom2019Robustness \cite{DBLP:conf/nips/CroceASDFCM021} and Wong2020Fast\cite{DBLP:conf/iclr/WongRK20}. All the models we use are implemented in the robustbench toolbox\footnote{https://github.com/RobustBench/robustbench} \cite{DBLP:conf/nips/CroceASDFCM021} and the models' parameters are also provided in \cite{DBLP:conf/nips/CroceASDFCM021}. These models showed classification accuracy of 83.60\%, 77.80\%, 77.40\%, 62.60\%, and 63.10\% on NIPS2017, respectively. For all these models, we chose their $L_{inf}$ version parameters due to we mainly extend $ L_{inf} $ attack in this paper. 

\textbf{Baselines:} We have two kind of baselines in this work. The classical methods including BIM \cite{DBLP:conf/iclr/KurakinGB17a}, PGD \cite{DBLP:conf/iclr/MadryMSTV18}, MIFGSM \cite{DBLP:conf/cvpr/DongLPS0HL18}, TIFGSM \cite{DBLP:conf/cvpr/DongPSZ19}, DIFGSM \cite{DBLP:conf/cvpr/XieZZBWRY19}, APGD \cite{DBLP:conf/icml/Croce020a} and Jitter \cite{DBLP:journals/corr/abs-2105-10304}. The experimental results of those methods are reproduced by the Torchattacks toolkit\footnote{https://github.com/Harry24k/adversarial-attacks-pytorch} with default settings. The another is the imperceptible methods, stAdv \cite{DBLP:conf/iclr/XiaoZ0HLS18}, Chroma-shift \cite{DBLP:conf/mm/AydinSKHT21} and the AdvFlow \cite{DBLP:conf/nips/DolatabadiEL20}. The codes used in here are provided by the corresponding authors.

All the experiments are conducted on a GPU server with 4 * Tesla A100 40GB GPU, 2 * Xeon Glod 6112 CPU, and RAM 512GB. 

% \vspace{-10pt}
\begin{table}
\centering
\setlength\tabcolsep{2pt}
\renewcommand{\arraystretch}{1.2}
\caption{Experimental results on the attack success rate of \textbf{un-targeted} attack on dataset NIPS2017 under $l_{inf} $ noise budget is 1, 2, and 4, respectively.}
\label{tab:NIPS2017_UN}
\resizebox{\linewidth}{!}{
\begin{tabular}{c|c|ccccccccc}
\hline
Epsilon            & Model        & BIM            & PGD   & MIFGSM & TIFGSM & DIFGSM & APGD  & Jitter & AdvFlow & AFLOW           \\ \hline
\multirow{4}{*}{1} & VGG-19       & 34.94          & 37.42 & 45.06  & 10.34  & 28.31  & 20.45 & 23.37  & 27.34   & \textbf{87.98}  \\
                   & ResNet-152   & 25.64          & 26.38 & 37.50   & 5.72   & 17.48  & 20.02 & 16.84  & 17.76   & \textbf{86.97}  \\
                   & MobileNetV2  & 41.8           & 43.17 & 51.25  & 11.50   & 30.98  & 22.21 & 21.30   & 29.96   & \textbf{93.97}  \\
                   & ShuffleNetV2 & 54.34          & 53.06 & 65.86  & 13.09  & 40.40   & 13.37 & 23.19  & 41.15   & \textbf{96.73}  \\ \hline
\multirow{4}{*}{2} & VGG-19       & 82.13          & 83.26 & 85.84  & 31.35  & 75.17  & 47.42 & 56.18  & 54.30    & \textbf{98.764} \\
                   & ResNet-152   & 68.22          & 69.81 & 75.85  & 17.48  & 55.72  & 49.58 & 50.53  & 41.31   & \textbf{99.26}  \\
                   & MobileNetV2  & 84.51          & 85.08 & 85.19  & 28.59  & 74.49  & 46.36 & 58.31  & 60.70    & \textbf{99.55}  \\
                   & ShuffleNetV2 & 89.90          & 90.33 & 91.32  & 32.72  & 75.68  & 23.61 & 51.21  & 74.22   & \textbf{100}    \\ \hline
\multirow{4}{*}{4} & VGG-19       & 98.20          & 98.76 & 98.43  & 71.01  & 97.53  & 56.29 & 83.60   & 82.03   & \textbf{99.66}  \\
                   & ResNet-152   & 93.75          & 95.34 & 95.13  & 50.32  & 93.01  & 66.95 & 84.42  & 77.43   & \textbf{99.79}  \\
                   & MobileNetV2  & 97.84          & 98.86 & 97.72  & 68.91  & 98.29  & 53.30  & 84.62  & 89.84   & \textbf{99.87}  \\
                   & ShuffleNetV2 & 98.44 & 98.86 & 98.72  & 67.99  & 97.30  & 25.75 & 71.55  & 92.97   & \textbf{100}    \\ \hline
\end{tabular}
}
\end{table}

% \vspace{-25pt}

\begin{table}[ht]
\caption{Experimental results on the attack success rate of \textbf{targeted} attack on dataset NIPS2017 under $l_{inf} $ noise budget is 1, 2, and 4, respectively.}
\label{tab:NIPS2017_T}
\centering

\setlength\tabcolsep{2pt}
\renewcommand{\arraystretch}{1.2}
\resizebox{\linewidth}{!}{
\begin{tabular}{c|c|ccccccccc}
\hline
Epsilon             & Model        & BIM   & PGD            & MIFGSM         & TIFGSM & DIFGSM & APGD  & Jitter & AdvFlow & AFLOW          \\ \hline
\multirow{4}{*}{1} & VGG-19       & 8.20   & 10.34          & 20.67          & 0.45   & 6.18   & 11.69 & 3.71   & 4.69    & \textbf{13.67} \\
                   & ResNet-152   & 6.78  & 9.64           & 20.13          & 0.21   & 2.97   & 11.23 & 1.91   & 3.51    & \textbf{17.58} \\
                   & MobileNetV2  & 14.35 & 19.13          & \textbf{39.41} & 0.68   & 9.34   & 20.16 & 3.42   & 5.34    & 39.06          \\
                   & ShuffleNetV2 & 20.34 & 20.91          & \textbf{41.68} & 0.43   & 5.97   & 21.64 & 5.55   & 7.16    & 41.41          \\ \hline
\multirow{4}{*}{2} & VGG-19       & 67.53 & 83.82          & 53.37          & 7.98   & 57.3   & 59.87 & 6.52   & 9.62    & \textbf{70.7}  \\
                   & ResNet-152   & 63.45 & 83.26          & 53.81          & 5.83   & 42.58  & 70.26 & 3.39   & 7.34    & \textbf{86.72} \\
                   & MobileNetV2  & 85.31 & \textbf{93.28} & 83.83          & 9.57   & 66.63  & 87.85 & 6.95   & 7.96    & 91.02          \\
                   & ShuffleNetV2 & 82.79 & 88.05          & 85.78          & 4.98   & 59.74  & 53.69 & 11.66  & 10.18   & \textbf{93.36} \\ \hline
\multirow{4}{*}{4} & VGG-19       & 95.51 & \textbf{99.44} & 76.07          & 56.52  & 96.07  & 98.65 & 9.44   & 23.56   & 98.83          \\
                   & ResNet-152   & 95.13 & 99.26          & 74.79          & 49.36  & 93.54  & 95.68 & 6.14   & 20.67   & \textbf{100}   \\
                   & MobileNetV2  & 98.75 & \textbf{99.89} & 94.31          & 71.07  & 98.18  & 98.48 & 11.16  & 24.25   & 99.61          \\
                   & ShuffleNetV2 & 99.29 & 99.72          & 98.01          & 52.20   & 98.29  & 99.36 & 18.63  & 29.31   & \textbf{100}   \\ \hline
\end{tabular}
}
\end{table}

% \vspace{-20pt}
\subsection{Quantitative Comparison with the Existing Methods}
In this subsection, we will evaluate the proposed AFLOW and the baselines BIM, PGD, MI-FGSM, TI-FGSM \cite{DBLP:conf/cvpr/DongPSZ19}, DI$^2$-FGSM \cite{DBLP:conf/cvpr/XieZZBWRY19}, APGD, Jitter, and AdvFlow in ASR on Caltech256 and Places365 dataset and the whole NIPS2017 dataset. We set the noise budget $ \epsilon $ of AFLOW and the baseline methods as 1, 2, and 4, respectively, for $ L_{inf} $ attack towards all the baseline methods under the non-target attack settings and the target attack settings. 

Table. \ref{tab:Caltech256}, \ref{tab:Places365}, \ref{tab:NIPS2017_UN}, and \ref{tab:NIPS2017_T} show the ASR on Caltech256, Places365 and NIPS2017, respectively. As can be seen, AFLOW can improve baseline methods' performance in most situations. Note that the proposed method can achieve an admirable attack success rate in a demanding perturbation budget, like $ \epsilon = 1 $. In contrast, other methods only get a relatively low attack success rate; take the non-target attack on NIPS2017 as an example. The BIM, PGD, MI-FGSM, TI-FGSM, DI$^2$-FGSM, APGD, and AdvFlow can only achieve 25.64\%, 26.38\%, 37.50\%, 5.72\%, 17.48\%, 20.02\%, 16.84\%, 17.76\% attack success rate on ResNet-152, respectively, vice versa, our AFLOW can achieve 86.79\% attack success rate. It is indicated that although these methods show fantastic attack performance in large noise budget settings, once we put a relatively extreme limit on the perturbation budget, these methods will lose their advantages completely and show dissatisfactory results. On the contrary, the AFLOW can attack the DNNs with smaller perturbations, in this setting, the adversarial examples generated by AFLOW are much less likely to be detected or denoised, so they are more threatening to DNNs and meaningful for exploring the existing DNNs' vulnerability and guiding the new DNNs' designing. 

% \vspace{-10pt}
\begin{table}[ht]
\caption{Experimental results on the attack success rate of \textbf{un-targeted} attack on dataset NIPS2017 to robust models under $l_{inf} $ noise budget is 1, 2, and 4, respectively.}
\label{tab:robust}
\centering
\setlength\tabcolsep{2pt}
\renewcommand{\arraystretch}{1.2}
\resizebox{\linewidth}{!}{
\begin{tabular}{c|c|ccccccccc}
\hline
Epsilon            & Methods                & BIM   & PGD   & MIFGSM & TIFGSM & DIFGSM & APGD  & Jitter         & \multicolumn{1}{l}{AdvFlow} & AFLOW          \\ \hline
\multirow{4}{*}{1} & Engstrom2019Robustness & 10.85 & 10.85 & 10.85  & 6.72   & 8.40   & 11.24 & 13.70          & 10.48                       & \textbf{15.21} \\
                   & Salman2020Do\_R18      & 12.36 & 12.36 & 12.36  & 8.78   & 10.62  & 12.52 & 15.37          & 11.78                       & \textbf{17.95} \\
                   & Salman2020Do\_R50      & 8.48  & 8.48  & 8.35   & 5.78   & 6.43   & 8.48  & 9.64           & 12.36                       & \textbf{10.35} \\
                   & Wong2020Fast           & 10.38 & 10.38 & 10.54  & 8.15   & 8.15   & 10.7  & 11.98          & 12.12                       & \textbf{12.02} \\ \hline
\multirow{4}{*}{2} & Engstrom2019Robustness & 23.77 & 24.03 & 23.26  & 15.50  & 19.51  & 24.55 & 26.74          & 27.31                       & \textbf{28.41} \\
                   & Salman2020Do\_R18      & 25.36 & 25.52 & 24.88  & 18.54  & 22.19  & 25.67 & 29.79          & 30.35                       & \textbf{31.52} \\
                   & Salman2020Do\_R50      & 18.12 & 18.12 & 17.74  & 12.21  & 14.91  & 18.25 & 20.69          & 26.92                       & \textbf{21.61} \\
                   & Wong2020Fast           & 20.61 & 20.45 & 21.41  & 15.81  & 17.41  & 22.36 & 25.24          & 28.08                       & \textbf{27.45} \\ \hline
\multirow{4}{*}{4} & Engstrom2019Robustness & 46.64 & 48.84 & 40.44  & 31.52  & 40.70  & 50.78 & 54.39          & 49.03                       & \textbf{55.30} \\
                   & Salman2020Do\_R18      & 46.91 & 46.59 & 43.74  & 36.29  & 43.42  & 47.23 & \textbf{53.25} & 47.94                       & 52.53          \\
                   & Salman2020Do\_R50      & 40.62 & 41.00 & 37.40  & 27.76  & 35.60  & 41.90 & 45.89          & 46.64                       & \textbf{48.42} \\
                   & Wong2020Fast           & 45.85 & 47.12 & 44.09  & 38.02  & 41.85  & 48.72 & \textbf{50.16} & 40.62                       & 41.50          \\ \hline
\end{tabular}
}
\end{table}

% \vspace{-20pt}
\subsection{Attack on Defense Models}
Next, we investigate the performance of the proposed method in attacking robust image classifiers. Thus we select some of the most recent defense techniques that are from the robustness toolbox as follows, Engstrom2019Robustness \cite{DBLP:conf/nips/CroceASDFCM021}, Salman2020Do\_R18 \cite{DBLP:conf/nips/SalmanIEKM20}, Salman2020Do\_R50 \cite{DBLP:conf/nips/SalmanIEKM20} and Wong2020Fast \cite{DBLP:conf/iclr/WongRK20}. We compare our proposed method with the baseline methods.

Following the results shown in Table. \ref{tab:robust}, we derive that AFLOW exhibits the best performance of all the baseline methods in terms of the attack success rate. Especially in a lower noise budget, like $\epsilon = 1$ or $\epsilon = 2$, the baseline methods range from 6.72\% to 27.31\% attack success rate on the Engstrom2019Robustness model. However, the AFLOW can obtain a higher performance range from 15.21\% to 28.41\%. It demonstrates the superiority of our method when attacking robust models.

\begin{table}[]
\caption{The detect results of AFLOW and the baselines.}
\label{tab:detect}
\centering
\setlength\tabcolsep{2pt}
\renewcommand{\arraystretch}{1.2}
\resizebox{\linewidth}{!}{
\begin{tabular}{c|c|cccc|cccc}
\hline
\multirow{2}{*}{Datasets} & \multirow{2}{*}{Methods} & \multicolumn{4}{c|}{AUROC (\%) $\uparrow$}                 & \multicolumn{4}{c}{Detection Acc. (\%) $\uparrow$}              \\ \cline{3-10} 
                          &                          & FGSM  & BIM   & AdvFlow        & AFLOW          & FGSM  & BIM   & AdvFlow        & AFLOW          \\ \hline
\multirow{3}{*}{CIFAR-10} & LID                      & 99.67 & 96.54 & 59.59          & \textbf{52.06} & 99.73 & 90.42 & \textbf{55.63} & 58.76          \\
                          & Mahalanobis              & 96.54 & 99.6  & 66.87          & \textbf{58.43} & 90.42 & 97.26 & 65.31          & \textbf{64.09} \\
                          & Res-Flow                 & 94.47 & 97.15 & 65.63          & \textbf{63.25} & 88.56 & 91.54 & 63.36          & \textbf{59.62} \\ \hline
\multirow{3}{*}{SVHN}     & LID                      & 97.86 & 90.55 & 62.57          & \textbf{62.13} & 93.34 & 82.6  & 59.21          & \textbf{57.65} \\
                          & Mahalanobis              & 99.61 & 97.14 & \textbf{64.84} & 65.36          & 98.62 & 92.49 & \textbf{61.57} & 62.56          \\
                          & Res-Flow                 & 99.07 & 99.42 & 65.68          & \textbf{64.98} & 95.92 & 96.99 & 63.73          & \textbf{62.69} \\ \hline
\end{tabular}
}
\end{table}

\subsection{Detectability}
Adversarial examples can be regarded as the data out of the distribution of the clean data, therefore we could check whether every example is adversarial or not. Thus, generating adversarial examples with high concealment means that they have the same or a similar distribution as the original data \cite{DBLP:conf/iclr/Ma0WEWSSHB18,DBLP:conf/nips/DolatabadiEL20}. To verify the crafted examples meet this rule, following the literature \cite{DBLP:conf/nips/DolatabadiEL20} and choose LID \cite{DBLP:conf/iclr/Ma0WEWSSHB18} , Mahalanobis \cite{DBLP:conf/nips/LeeLLS18}, and Res-Flow \cite{DBLP:conf/cvpr/ZisselmanT20} adversarial attack detectors to evaluate the performance of the AFLOW. For comparison, we choose FGSM \cite{DBLP:journals/corr/GoodfellowSS14}, BIM \cite{DBLP:conf/iclr/KurakinGB17a}, and AdvFlow \cite{DBLP:conf/nips/DolatabadiEL20} as the baseline methods. The detection results are shown in Table. \ref{tab:detect}, including the area under the receiver operating characteristic curve (AUROC) and the detection accuracy. From Table. \ref{tab:detect}, we can find that these adversarial detectors find it hard to detect the evil examples built by AFLOW in contrast to the baselines in most cases. The empirical results precisely demonstrate the superiority of our method, which generates adversarial examples closer to the original clean images' distribution than other methods, and the optimized adversarial perturbations have better hiding ability. The classifier is ResNet-34 and the code used in this experiment is modified from deep\_Mahalanobis\_detector\footnote{https://github.com/pokaxpoka/deep\_Mahalanobis\_detector} and Residual-Flow\footnote{https://github.com/EvZissel/Residual-Flow}, respectively.

% \vspace{-0.5 cm}
\begin{table}[htp]
\centering
\setlength\tabcolsep{2pt}
\renewcommand{\arraystretch}{1.2}

\caption{Various perceptual distances were calculated on fooled examples by BIM, PGD, MI-FGSM, TI-FGSM, DI$^2$-FGSM, APGD, Jitter, stAdv, Chroma-Shift, AdvFlow and the proposed AFLOW on NIPS2017.}
\resizebox{\linewidth}{!}{
\begin{tabular}{c|ccccccccccc}
\hline
Metrics             & BIM     & PGD     & MI-FGSM & TI-FGSM & DI-FGSM & APGD    & Jitter  & \multicolumn{1}{l}{stAdv} & \multicolumn{1}{l}{Chroma-Shift} & \multicolumn{1}{l}{AdvFlow} & AFLOW            \\
SSIM $\uparrow$     & 0.9496  & 0.8905  & 0.9446  & 0.9193  & 0.9186  & 0.8727  & 0.9094  & 0.9565                    & 0.9760                           & 0.9863                      & \textbf{0.9952}  \\
PSNR $\uparrow$     & 36.6813 & 33.1693 & 36.2556 & 33.5426 & 34.6539 & 32.6917 & 33.5590 & 31.0612                   & 35.1582                          & 34.1804                     & \textbf{36.7962} \\
UQI $\uparrow$      & 0.9821  & 0.9768  & 0.9837  & 0.9653  & 0.9839  & 0.9812  & 0.9828  & 11.9378                   & 7.6892                           & 7.8021                      & \textbf{0.9844}  \\
SCC $\uparrow$      & 0.7277  & 0.6085  & 0.7068  & 0.8145  & 0.6798  & 0.5919  & 0.6423  & 0.7109                    & 0.8496                           & 0.9041                      & \textbf{0.9611}  \\
VIFP $\uparrow$     & 0.6516  & 0.5393  & 0.6522  & 0.5551  & 0.5838  & 0.5172  & 0.5897  & 0.5614                    & 0.7297                           & 0.8027                      & \textbf{0.8649}  \\
L$_2$  $\downarrow$ & 56.8518 & 84.3255 & 59.7074 & 81.5976 & 71.7985 & 89.9959 & 81.4444 & 0.9976                    & 0.9970                           & \textbf{0.9831}             & 56.4112          \\
LPIPS $\downarrow$  & 0.1490  & 0.2133  & 0.1580  & 0.1646  & 0.1993  & 0.2391  & 0.1962  & 0.1338                    & 0.0203                           & 0.0226                      & \textbf{0.0101}  \\
DISTS $\downarrow$  & 0.1022  & 0.1383  & 0.1054  & 0.1391  & 0.1398  & 0.1545  & 0.1272  & 0.1360                    & 0.0246                           & 0.0263                      & \textbf{0.0204}  \\ \hline
\end{tabular}
}
\label{tab:perceptual}
\end{table}

\subsection{Evaluation of Image Similarity}
In this paper, we follow the work in \cite{DBLP:conf/mm/AydinSKHT21} using the following perceptual metrics to evaluate the adversarial examples generated by our method: Learned Perceptual Image Patch Similarity (LPIPS) metric \cite{DBLP:conf/cvpr/ZhangIESW18}, and Deep Image Structure and Texture Similarity (DISTS) index \cite{DBLP:journals/pami/DingMWS22}. LPIPS is a technique that measures the Euclidean distance of deep representations (i.e., VGG network \cite{DBLP:journals/corr/SimonyanZ14a}) calibrated by human perception. Moreover, we also use the Structure Similarity Index Measure (SSIM) \cite{DBLP:journals/tip/WangBSS04} to assess the generated images' qualities concerning luminance, contrast, and structure. Next, we calculate the Average $L_2$ norm. Finally, we use other metrics like Universal Image Quality Index (UQI) \cite{DBLP:journals/spl/WangB02}. Spatial Correlation Coefficient (SCC) \cite{scc}, and Pixel Based Visual Information Fidelity (VIFP) \cite{DBLP:conf/icassp/SheikhB04} to assess the adversarial examples' image quality. The main toolkits we used in the experiments of this part are IQA\_pytorch\footnote{https://www.cnpython.com/pypi/iqa-pytorch} and sewar\footnote{https://github.com/andrewekhalel/sewar}.

The generated images' quality results can be seen in Table. \ref{tab:perceptual}, which indicated that the proposed method has the lowest LPIPS, and DISTS perceptual loss (the lower is better), are 0.0101 and 0.0204, respectively, and has the highest SSIM, PSNR, UQI, SCC and VIFP (the higher is better), achieving 0.9952, 36.7962, 0.9844, 0.9611, and 0.8649, respectively, in comparison to the baselines on NIPS2017 dataset. The results show that the proposed method is superior to the existing attack methods.

\begin{figure}[htp]
      \centering
      \includegraphics[width=\textwidth]{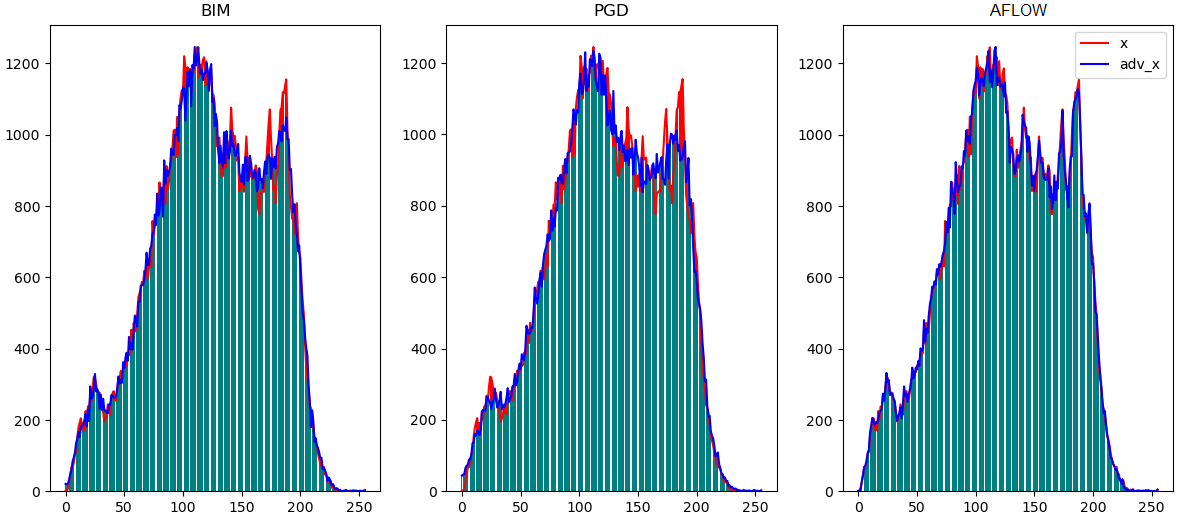}
      % \vspace{-0.9 cm}
      \caption{The gray histogram comparison among baselines and our method between clean example and adversarial example, with the red line represent the benign example and the blue line indicate the corresponding adversarial one.}
      \label{fig:his}
\end{figure}

In addition, we draw the gray histogram of the adversarial example generated by BIM, PGD, and our method in Fig. \ref{fig:his} to show the modification of the original image. The horizontal axis represents the pixel's value, and the vertical axis represents the number of pixels corresponding to each pixel value. From Fig. \ref{fig:his}, we can see that the adversarial examples generated by AFLOW are more similar to the original image, and the distribution of the number of pixel values is almost the same as the original image. While the baseline methods BIM and PGD change the original image a lot, resulting in a significant difference in the distribution of the number of pixel values.

\begin{figure}[htbp]
      \centering
      \includegraphics[width=\textwidth]{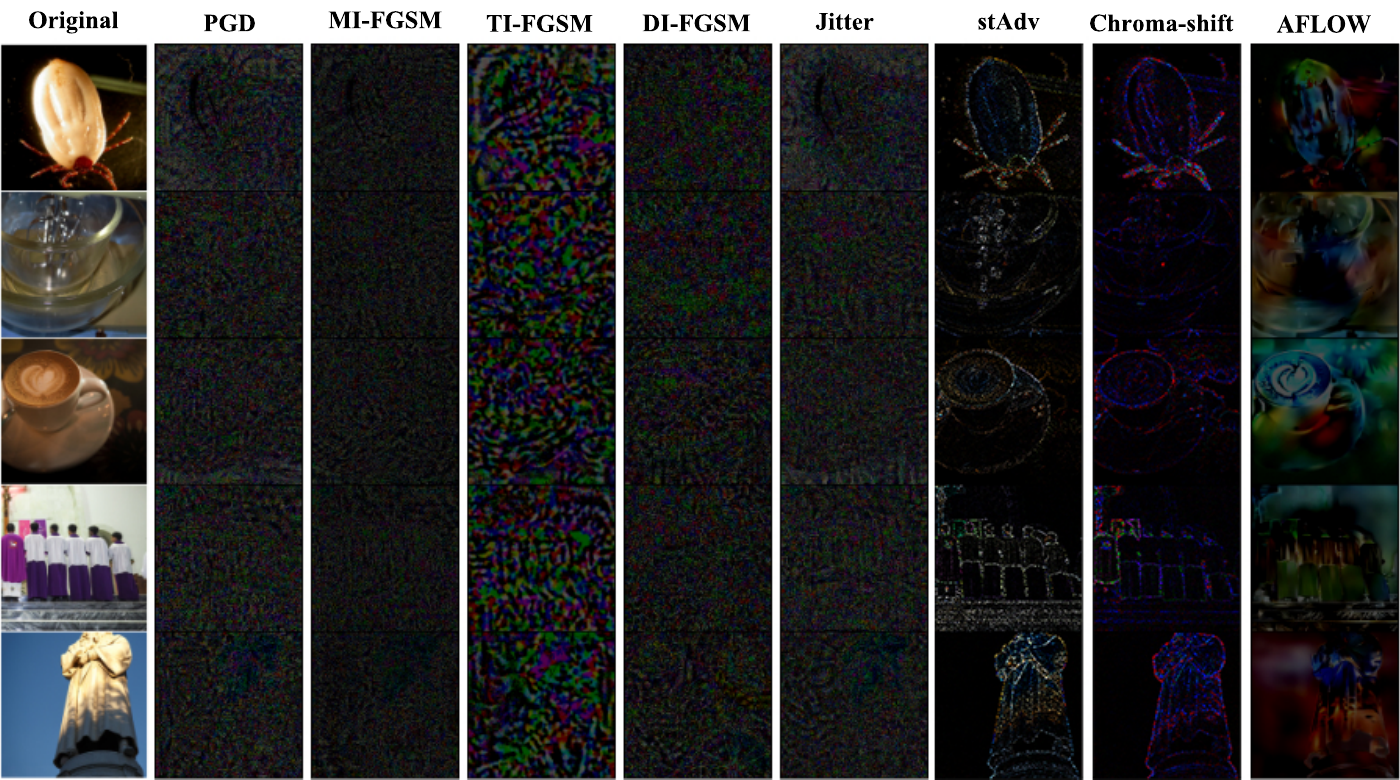}
      \caption{Adversarial examples and their corresponding perturbations. The first column is the benign examples, and the followings are the adversarial noise of PGD, MI-FGSM, TI-FGSM, DI$^2$-FGSM, Jitter, stAdv, Chroma-shift, and our method, respectively.}
      \label{fig:noise}
\end{figure}

To better observe the difference between the adversarial examples generated by our method and the baselines from the visual aspect, we also draw the adversarial perturbation generated on NIPS2107 by baselines and the proposed method in Fig. \ref{fig:noise}, the target model is pre-trained ResNet-152. The first column is the benign examples, and the following are the adversarial noise of PGD, MI-FGSM, TI-FGSM, DI$^2$-FGSM, Jitter, stAdv, Chroma-shift and our method, respectively. Noted that, for better observation, we magnified the noise by a factor of 10. From Fig. \ref{fig:noise}, we can clearly observe that baseline methods distort the image without ordering. In contrast, the adversarial examples generated by our method are focused on the target object, and its noise contains more semantic information, and they are similar to the original clean image and are more imperceptible to human eyes.

\section{Conclusions}
\label{Sec:conclusion}
In this paper, we present a novel study on the adversarial attack in a rigorous noise-limited scenario, explicitly focusing on the CV task. To ensure the perturbation is unnoticeable, we generate adversarial examples by directly disturbing the images' hidden representation rather than noise-adding. The proposed method, called AFLOW, based on Normalize Flow model, has succeeded in improving attack ability and enhancing the imperceptibility of the generated adversarial noise. Extensive experimental results show the proposed AFLOW can generate adversarial examples with high attack ability, admirable invisibility, and excellent image quality. This work may be a starting point for future research on sufficiently evaluating the existing DNNs' vulnerability. Where several issues could be further investigated, including further helping consolidate the existing DNNs and designing new robust DNN models.

\section{Acknowledgments}
This work is supported in part by Yunnan Province Education Department Foundation under Grant No.2022j0008, in part by the National Natural Science Foundation of China under Grant 62162067 and 62101480, Research and Application of Object Detection based on Artificial Intelligence, in part by the Yunnan Province expert workstations under Grant 202205AF150145.

\bibliographystyle{splncs04}
\bibliography{ref}

\end{document}